\documentclass[10pt,twocolumn,letterpaper]{article}

\usepackage{iccv}
\usepackage{times}
\usepackage{epsfig}
\usepackage{graphicx}
\usepackage{amsmath}
\usepackage{amssymb}
\usepackage{amsmath,amssymb}
\usepackage{float}

\usepackage[pagebackref=true,breaklinks=true,letterpaper=true,colorlinks,bookmarks=false]{hyperref}

\iccvfinalcopy 

\ificcvfinal\fi

\begin{document}

\title{GBSD: Generative Bokeh with Stage Diffusion}
%

\author{Jieren Deng\thanks{The work was done when the $1^{st}$ author was an intern at Baidu Research.}\\
University of Connecticut\\
\and
Xin Zhou\\
Baidu Research USA\\
\and
Hao Tian\\
Baidu Research USA\\
\and
Zhihong Pan\\
Baidu Research USA\\
\and
Derek Aguiar\\
University of Connecticut\\
}

\maketitle

\begin{abstract}
   The bokeh effect is an artistic technique that blurs out-of-focus areas in a photograph and has gained interest due to recent developments in text-to-image synthesis and the ubiquity of smartphone cameras and photo sharing apps. 
   Prior work on rendering bokeh effects have focused on \textit{post hoc} image manipulation to produce similar blurring effects in existing photographs using classical computer graphics or neural rendering techniques, but have either depth discontinuity artifacts or are restricted to reproducing bokeh effects that are present in the training data.
   More recent diffusion based models can synthesize images with an artistic style, but either require the generation of high-dimensional masks, expensive fine-tuning, or affect global image characteristics.
   In this paper, we present \texttt{GBSD}, the first generative text-to-image model that synthesizes photorealistic images with a bokeh style. 
   Motivated by how image synthesis occurs progressively in diffusion models, our approach combines latent diffusion models with a 2-stage conditioning algorithm to render bokeh effects on semantically defined objects.
   Since we can focus the effect on objects, this \textit{semantic bokeh} effect is more versatile than classical rendering techniques.
   We evaluate \texttt{GBSD} both quantitatively and qualitatively and demonstrate its ability to be applied in both text-to-image and image-to-image settings. 
\end{abstract}


\section{Introduction}

The bokeh effect refers to an artistic styling in photography that creates an out-of-focus blurring in areas of a photograph.
Photographers have traditionally achieved the bokeh blurring effect by widening the lens aperture or through lens aberrations.
The ubiquity of portable smartphone cameras and photo sharing applications has generated increased interest in image manipulation in general and bokeh synthesis specifically~\cite{ignatov2019aim,ignatov2020aim}.
Prior work is primary concerned with \textit{post hoc} image manipulation to produce similar blurring effects in existing photographs using classical computer graphics~\cite{lee2010real,wu2013rendering,yu2010real} or neural rendering techniques~\cite{pengBokehMeWhenNeural2022,nalbach2017deep,xu2018rendering,ignatov2020rendering,dutta2021stacked,wang2018deeplens,xiao2018deepfocus}. 
Neural rendering resolves the depth discontinuity artifacts present in classical techniques, but are generally restricted to reproducing bokeh effects that are present in the training data~\cite{pengBokehMeWhenNeural2022}  or typical of classical techniques (e.g., a bokeh ball effect). 
Recent work achieves arbitrary blur sizes and shapes, but requires high-dimensional maps that are difficult to generate~\cite{pengBokehMeWhenNeural2022}. 
Moreover, all prior methods assume there exists an input image to be manipulated, that is, they are not fully generative models.


Recently, diffusion models~\cite{NEURIPS2021_49ad23d1} have demonstrated an ability to generate photorealistic images given a text prompt~\cite{ramesh2022hierarchical,saharia2022photorealistic,yu2022scaling}.
The artistic characteristics of synthesized images can be controlled through \textit{post hoc} image manipulation or within the generative process.
Image editing tasks like image inpainting~\cite{lugmayr2022repaint,saharia2022image,peng2021generating,zhao2020uctgan} are typically cast as image-to-image translations or require a user specified mask (and thus presuppose an image) to define a location in the image to edit.
Synthesizing images with a specified artistic styling can be achieved by conditioning on class labels or text descriptions~\cite{saharia2022photorealistic,yu2022scaling,ramesh2022hierarchical}.
However, these control signals typically affect global image characteristics like artistic style and bokeh effects have not be achieved in diffusion models.

\begin{figure*}[h] 
\centering
\includegraphics[width=1.8\columnwidth]{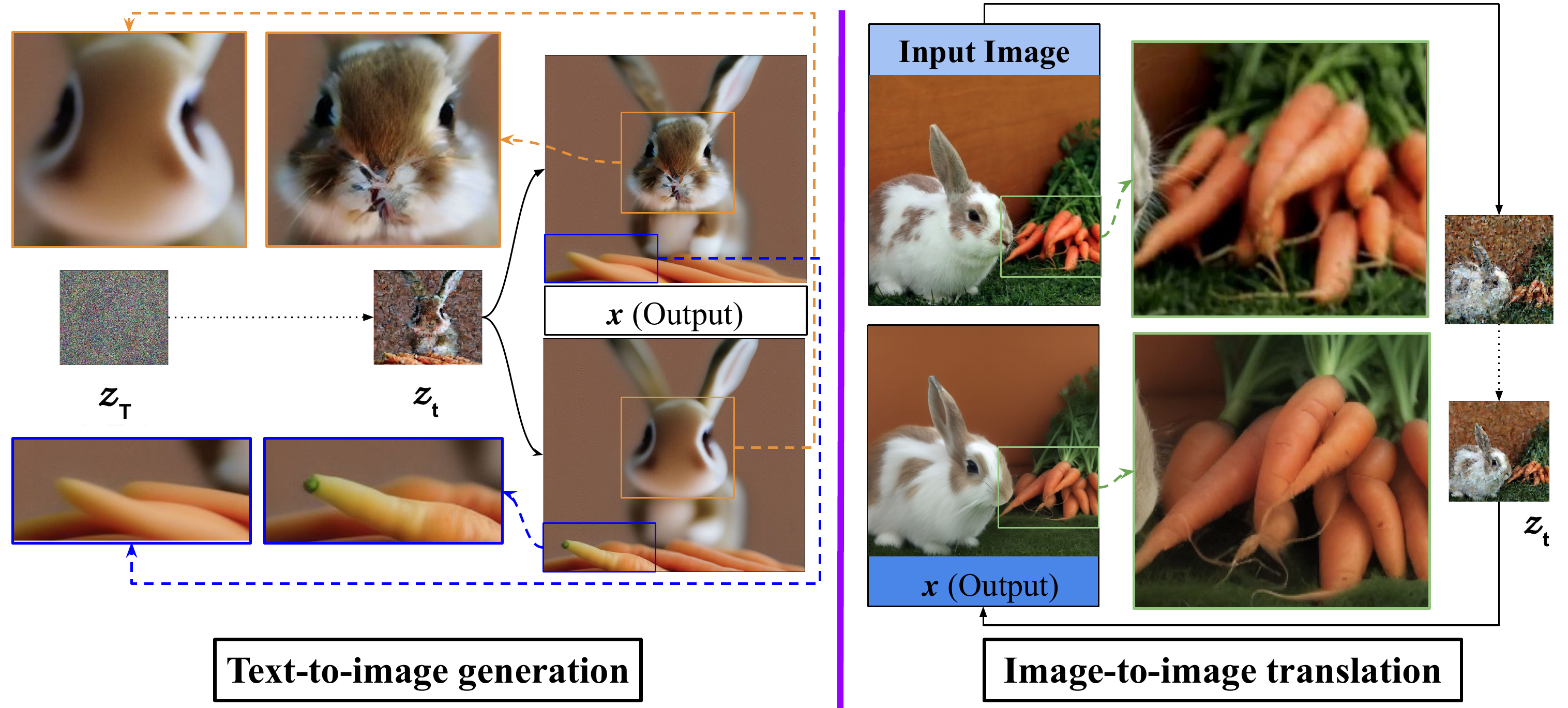}
\caption{\textbf{An illustration of stage diffusion for text-to-image and image-to-image generation with a bokeh style.} 
A bokeh style image is generated by a two-stage semantic conditioning algorithm.
The first stage (from $z_{T}$ to $z_{t}$) generates the \textit{global layout} of the image (e.g., shape and color) while stage two (from $z_{t}$ to $x$) \textit{focuses} detail and bokeh effects through semantic conditioning.
In the text-to-image example (left), the stage 1 prompt was \textit{``A cute baby bunny standing on top of a pile of baby carrots under a spot light"} and different prompts in stage 2 that focus either the carrots (bottom) or the bunny (top).
In the image-to-image example (right), we use the prompt \textit{``A cute rabbit stands with carrots with green leaf"} in stage 1 and \textit{``carrots with green leaf"} in stage 2. 
The generated image demonstrates the previously blurry carrot coming into focus, revealing more clear and distinct textures, while creating a bokeh effect for the rabbit.
}
\label{Fig:lambda}
\end{figure*}

In this work, we present \texttt{GBSD}, the first generative text-to-image model capable of synthesizing photorealistic images with a bokeh style.
Motivated by how image synthesis occurs progressively in diffusion models, that is, image layout, shape, and color are generated before enhancing details~\cite{10.5555/3495724.3496298}, our approach combines latent diffusion models with a 2-stage conditioning algorithm to render bokeh effects on semantically defined objects (Fig.~\ref{Fig:lambda}).
The two stages apply different text conditioning to the latent diffusion network;
the first (global layout) stage generates the structure of the image (e.g., the shape and color of objects) and the second (focus) stage simultaneously focuses detail generation and bokeh on different objects.
Since we can focus the effect on objects, this \textit{semantic bokeh} effect is more versatile than classical rendering techniques.
Due to the simplicity of our conditioning algorithm, \texttt{GBSD} does not require the specification of a high-dimensional mask or expensive retraining. 

Our work makes the following contributions:
\begin{itemize}
    \item We present a new problem, semantic bokeh, whose goal is to apply the bokeh blur effect to semantically distinct objects in an image. 
    \item We propose \texttt{GBSD}, a generative photorealistic image synthesis method based on latent stage diffusion. 
    \texttt{GBSD} is the first diffusion model capable of synthesizing photorealistic bokeh stylized images and can be applied in both text-to-image (Fig.~\ref{Fig:lambda}, left) and image-to-image (Fig.~\ref{Fig:lambda}, right) settings.
    \item We evaluate our bokeh stage diffusion model both quantitatively and qualitatively by varying stage time and diffusion prompts for text-to-image and image-to-image tasks.
\end{itemize}

\section{Related Work}

\paragraph{Bokeh Synthesis.}
Methods for rendering bokeh include classical techniques that are typically inefficient (e.g., ray tracing~\cite{lee2010real,wu2013rendering}) or require highly structure prior information like 3D scenes or depth maps (e.g., image space blurring or defocusing~\cite{yu2010real,yang2016virtual,hach2015cinematic,bertalmio2004real,barron2015fast}).
Taking advantage of advances in computer vision, subsequent bokeh rendering methods improved on classical techniques through the integration with image segmentation~\cite{shen2016automatic,shen2016deep}, depth perception~\cite{busam2019sterefo,luoBokehRenderingDefocus2020,peng2021interactive,xian2021ranking,zhang2019synthetic}, or both~\cite{wadhwa2018synthetic}.
More recently, neural bokeh synthesis techniques were developed to address the depth discontinuity artifacts around boundaries and lack of scalability of classical methods.
Neural translation of in-focus to bokeh images using depth maps are accurate~\cite{nalbach2017deep,xiao2018deepfocus}, but generating high-quality depth maps is unrealistic in most scenarios.
Methods that performed depth prediction or employed encoder-decoder architectures followed~\cite{wang2018deeplens,pengBokehMeWhenNeural2022,xu2018rendering,ignatov2020rendering,dutta2021stacked,ignatov2019aim,ignatov2020aim}, but are, by construction, limited to reproducing bokeh effects similar to the data used to train them or require the generation of high-dimensional maps. 
Additionally, while some prior work employ generative models~\cite{dutta2021stacked,ignatov2019aim,ignatov2020aim,qianBGGANBokehGlassGenerative2020}, all prior bokeh synthesis methods aim for image-to-image translation, assuming an input image exists to be manipulated.

\paragraph{Deep Generative Models.}
Early architectures for synthesizing images from text relied on generative adversarial networks (GANs)~\cite{gan} and variational autoencoders (VAE)~\cite{Kingma2014}.
GANs are capable of generating high-quality and high-resolution images, but are difficult to optimize~\cite{10.5555/3295222.3295327,Mescheder2018OnTC} and can drop regions of the data distribution~\cite{metz2017unrolled}.
In contrast with GANs, VAEs~\cite{Kingma2014} can efficiently synthesize high-resolution images but typically generate lower quality images~\cite{https://doi.org/10.48550/arxiv.2011.10650}. 
More recently, probabilistic diffusion models (DM)~\cite{10.5555/3495724.3496298,pmlr-v37-sohl-dickstein15,song2021scorebased}, which are based on an iterative denoising process~\cite{NEURIPS2021_49ad23d1}, have demonstrated state-of-the-art results across a variety of applications including text-to-image generation~\cite{ramesh2022hierarchical}, natural language generation~\cite{li2022diffusionlm}, time series prediction~\cite{lopezalcaraz2023diffusionbased}, medical image~\cite{10.1007/978-3-031-18576-2_12}, audio generation~\cite{kong2021diffwave}, adversarial machine learning~\cite{xiao2023densepure} and privacy-preserving machine learning~\cite{dockhorn2022differentially}.

\paragraph{Text Driven Image Generation and Editing Using DM.} 
A primary application of diffusion models is image synthesis and manipulation based on conditioning text, which includes text-to-image~\cite{mansimov2015generating,reed2016generative} and image-to-image generation. 
The denoising task can be conditioned by text prompts~\cite{Kim_2022_CVPR} in image space (GLIDE~\cite{pmlr-v162-nichol22a} and Imagen~\cite{saharia2022photorealistic} or in latent space, which includes DALL$\cdot$E 2~\cite{ramesh2022hierarchical}, latent diffusion models (LDM)~\cite{Rombach_2022_CVPR_ldm}) and vector quantized diffusion~\cite{https://doi.org/10.48550/arxiv.2111.14822}. 
To improve computational efficiency, it is common practice to train a diffusion model using low-resolution images or latent variables, which are then processed by super-resolution diffusion models~\cite{ho2021cascaded} or latent-to-image decoders~\cite{NEURIPS2021_682e0e79}.

\paragraph{Variable Text Prompt Conditioning.}
Adjusting the text conditioning during the denoising process has been considered in the image manipulation context.
Imagic uses a 3-step process to linearly interpolate between a target and optimized textual embeddings based on a reference image and text prompt~\cite{kawar2022imagic}.
However, Imagic requires the optimization of text embeddings based on pre-trained diffusion models, followed by diffusion model fine-tuning using the optimized text embeddings for each text prompt input.
The eDiff-I method changes the text prompt after a fixed percentage of denoising steps~\cite{https://doi.org/10.48550/arxiv.2211.01324}, though the focus is on evaluating the strength of text conditioning and denoising efficacy under different noise levels, rather than exploring how prompt switching affects the photographic properties of the generated image.
In contrast with prior work, we investigate how to split a continuous denoising process into two stages and leverage the prompt in stage 2 to simultaneously sharpen a target object while introducing a bokeh effect in others for both text-to-image generation and image-to-image generation.
Further, our model does not require the generation of a mask or expensive fine-tuning.

\section{Methods}
\subsection{Diffusion Model Preliminaries}
Probabilistic diffusion models estimate the data distribution $p(x)$ by denoising a normally distributed random variable with an input image $x \in \mathbb{R}^{H \times W \times 3}$ in RGB space. 
The denoising process is represented as the reverse of a length $T$ Markovian diffusion process~\cite{Rombach_2022_CVPR_ldm}, with the best performing image generation models using a weighted variational lower bound on $p(x)$~\cite{NEURIPS2021_49ad23d1,10.5555/3495724.3496298}. 
Let $x_t$ be a noisy version of the input $x$ and $\epsilon_{\theta}(x_t,t)$ be a denoising autoencoder with input $x_t$ at step $t$.
The diffusion process can be represented as denoising autoencoders $\{\epsilon_{\theta}(x_t,t)\}_{t=1}^T$, which are trained to predict a denoised variant of the input $x_t$. 
A simplified objective can be formulated as:
\begin{equation*}
\textit{L}_{DM} = \mathbb{E}_{x,\epsilon \sim \mathcal{N}(0,1), t} \left[ ||\epsilon-\epsilon_{\theta}(x_t,t)||^{2}_{2} \right]
\end{equation*}

Latent diffusion models (LDMs) leverage trained perceptual compression models $\varepsilon$ and $D$, where $\varepsilon(x)$ is an encoder for input $x$ to latent space $z$ and $D(z)$ decodes $z$ from the latent space back to image space producing $\bar x$~\cite{Rombach_2022_CVPR_ldm}. 
The model also uses a textual conditioning prompt $y$, which can be projected into an embedded representation through a parameterized domain-specific expert $\tau_{\theta}$. 
A new objective $L$ using this latent space representation can be formulated as:
\begin{equation*}
\textit{L}_{LDM} = \mathbb{E}_{z, y, \epsilon \sim \mathcal{N}(0,1), t} \left[ ||\epsilon-\epsilon_{\theta}(z_t,t,\tau_{\theta}(y))||^{2}_{2} \right]
\end{equation*}


Most diffusion models denoise with consistent and continued conditioning. 
To achieve a bokeh effect in synthesized images, we design a two-stage diffusion with a related, but distinct conditioning mechanism named \textit{stage diffusion}.
\begin{figure}[h] 
\centering
\includegraphics[width=1.05\columnwidth]{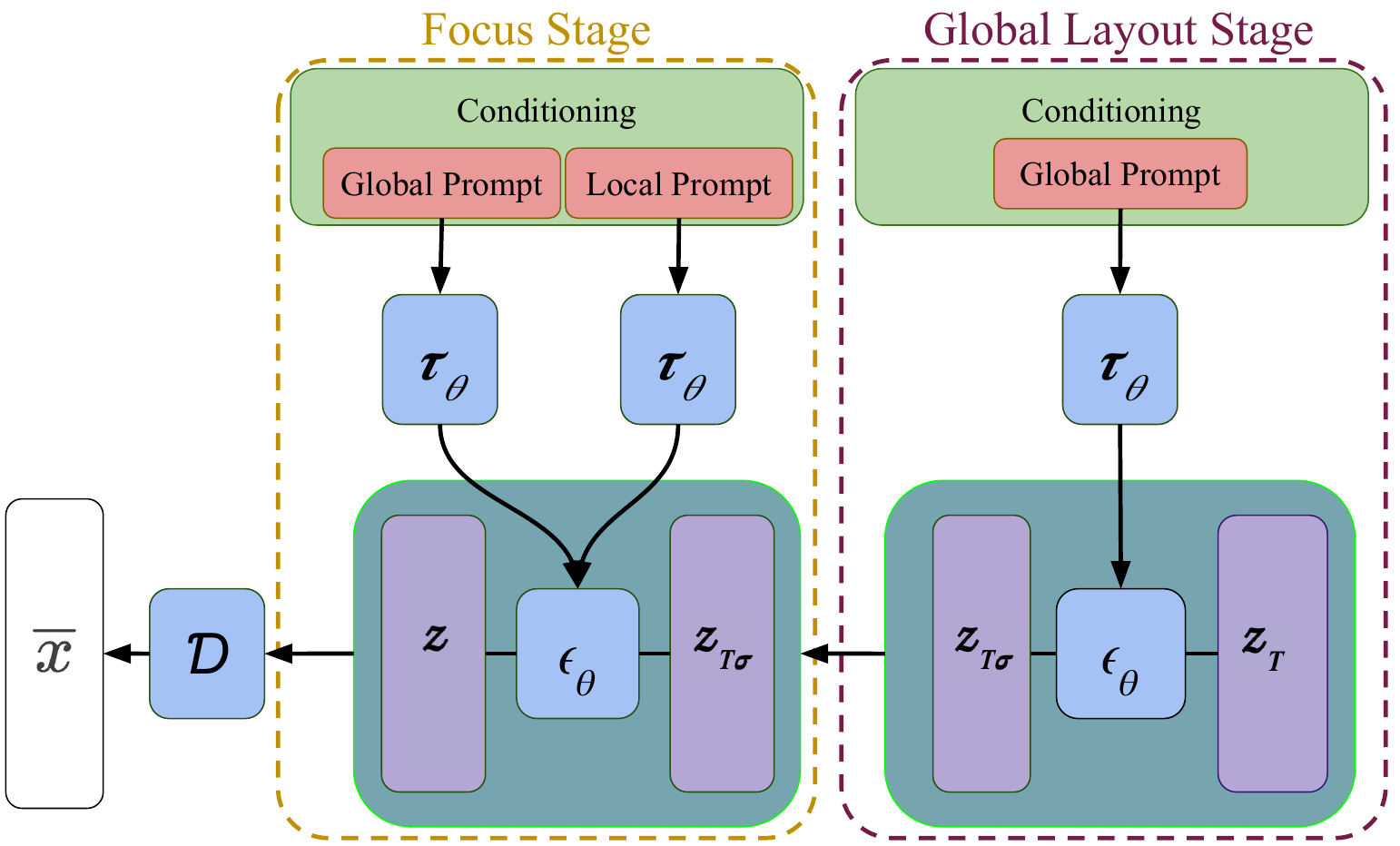}
\caption{\textbf{The generative bokeh with stage diffusion (\texttt{GBSD}) architecture.} 
For brevity, we represent the diffusion process for each stage by a single representative denoising autoencoder.}
\label{Fig:stage_diffusion}
\end{figure}
\subsection{Stage Diffusion}

The goal of stage diffusion is to synthesize an image that simultaneously focuses on a target object while producing a bokeh effect on others in both text-to-image and image-to-image generative scenarios.
Our stage diffusion method leverages LDMs~\cite{Rombach_2022_CVPR_ldm}, which provide a consistent text conditioning signal in each denoising autoencoder during image synthesis, and the progressive manner in which diffusion models synthesize images (i.e., generating image layout, shape, and color before enhancing details~\cite{10.5555/3495724.3496298}).
We implement stage diffusion by decomposing the diffusion process into two distinct stages: a \textit{global layout} stage and a \textit{focus} stage (Fig.~\ref{Fig:stage_diffusion}). 
The global layout stage generates the structure of the image (e.g., layout, shape, and color), whereas the focus stage outputs a final bokeh-styled image.

\subsubsection{Global Layout Stage}

In the global layout stage, we process a global prompt that completely describes the image to synthesize, $y_{global}$, through a domain-specific expert, $\tau_{\theta}$ to obtain its corresponding textual embedding $\tau_{\theta}(y_{global})$~\cite{pmlr-v139-radford21a}.
The model uses $\tau_{\theta}(y_{global})$ as a textual input for conditioning the diffusion process and employs a hyperparameter $\sigma \in (0,1)$ to regulate the number of denoising steps during  global layout stage. 
The denoising steps for a given global layout stage are set as the product of the total number of denoising steps $T$ and $\sigma$. 
If we consider the global layout stage as a function of $g(\cdot)$ with inputs, $\tau_{\theta}(y_{global})$, $z_{T}$ and $\sigma$, we represent the output of global layout stage $z_{\sigma\cdot T}$, as:
\begin{equation*}
z_{\sigma\cdot T} = g\left( \tau_{\theta}(y_{global}), z_{T},\sigma \right)
 \end{equation*}
The output of the global layout stage is an intermediate image with a stable and consistent structure for synthesized objects. 

\subsubsection{Focus Stage}
In order to simultaneously sharpen details on some objects while producing a bokeh effect in others, we pass a local prompt $y_{local}$ to the text encoder~\cite{pmlr-v139-radford21a}, which produces the textual embedding, $\tau_{\theta}(y_{local})$; the local prompt $y_{local}$ should be semantically related to the focused object. 
We linearly interpolate $\tau_{\theta}(y_{global})$ from the global layout stage and $\tau_{\theta}(y_{local})$ with a hyperparameter $\alpha$ as:
\begin{equation*}
\Bar{\textbf{\textit{e}}} = \left( \tau_{\theta}(y_{local}) + \alpha \times \tau_{\theta}(y_{global}) \right)/(1+\alpha)
 \end{equation*}
The resulting $\Bar{\textbf{\textit{e}}}$ represents the textual conditioning in focus stage. 
The parameter $\alpha$ balances the global and local information in the focus stage, impacting the final generated image. 
The denoising diffusion objective $L$ in focus stage is represented as:
\begin{equation*}
\textit{L} = \mathbb{E}_{z, y, \epsilon \sim \mathcal{N}(0,1), t} \left[ ||\epsilon-\epsilon_{\theta}(z_{\sigma\cdot T},t,\Bar{\textbf{\textit{e}}})||^{2}_{2} \right]
\end{equation*}
where $t = 1\dots(\sigma\times T)$.
Compared with other similar image synthesis or manipulation methods~\cite{kawar2022imagic,nalbach2017deep,xiao2018deepfocus}, stage diffusion does not require expensive fine-tuning or high-dimensional mask generation.

\subsection{Image Generation}
Stage diffusion is capable of both text-to-image and image-to-image synthesis.
Text-to-image generation does not require the initial diffusion process step. 
Instead, an arbitrary text prompt $y_{global}$ is input into the model and the initial random input is created by concatenating a sample from a Gaussian random variable with $\tau_{\theta}(y_{global})$ to create $z_T$.
For image-to-image generation, we add diffusion noise based on the desired amount of image perturbation.
In the global layout stage, a related or matched prompt is used as the global prompt to ensure the preservation of objects and prevent image distortion. 
This allows the model to generate outputs that are consistent with the given prompt, ensuring high-quality results.

\begin{figure*}[htbp]
\centering
\includegraphics[width=2.1\columnwidth]{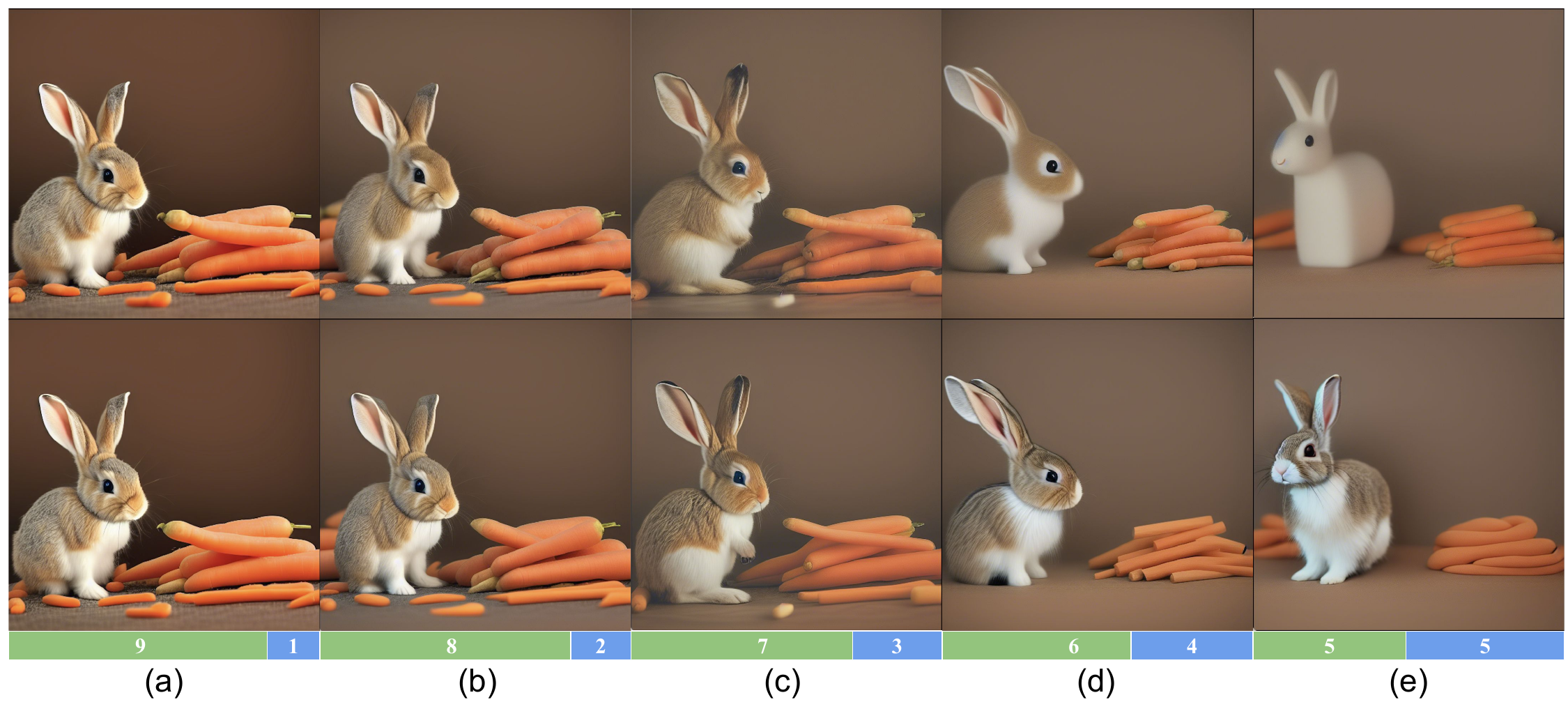}
\caption{\textbf{Comparing image generation by adjusting $\sigma$. }
The proportion of denoising steps allocated to the global layout stage was set to \{0.9,0.8,0.7,0.6,0.5\} and the overall number of denoising steps and conditioning per stage was fixed.
Both experiments (top and bottom rows) shared the same global prompt \textit{``a cute baby bunny standing on top of a pile of baby carrots under a spot light"}. 
The top row was given a local prompt of \textit{``a pile of carrots"} and the bottom row was given \textit{``a cute rabbit"}. 
The proportion of denoising steps spent in the global layout (stage 1) and focus (stage 2) stages is given in the green and blue boxes, respectively. 
The \textit{rabbit} features in the top experiment degenerated as stage 1 was shortened; similarly, the \textit{carrots} features in the bottom experiment exhibited a similar trend. 
}
\label{Fig:stage_comparison}
\end{figure*}

\section{Results}
We implement our generative bokeh with stage diffusion (\texttt{GBSD}) method both qualitatively and quantitatively and in text-to-image and image-to-image scenarios. 
All experiments are conducted using NVIDIA V100 GPUs with an image output size of $512\times512$, a batch size of $4$ (i.e., the number of samples to generate for each prompt), and a fixed random seed of $42$.
The code repository leveraged the officially released version of LDM (Stable Diffusion v1-4)~\cite{Rombach_2022_CVPR_ldm,stablediffhug}. 
For all experiments, the number of denoising diffusion implicit model (DDIM) sampling steps was set to 50, the number of timesteps was set to 1000, and the scale was set to 15 for all groups~\cite{song2021denoising}. 
Additional results are provided in the appendix.

\subsection{Evaluation Measures}

Based on a recent evaluation of focus measure operators~\cite{Pertuz2013AnalysisOF}, we consider variance of Laplacian~\cite{blur1,blur2} score and Brenner score~\cite{Zahniser} for quantitative evaluations.

\subsubsection{Variance of Laplacian}

The variance of Laplacian (VoL) is a measure that uses the variance of the image Laplacian for an evaluation of blur or focus~\cite{blur1,blur2} and is calculated by
\begin{equation*}
\sum_{(i, j) \in \Omega(x, y)} \left(\Delta I(i, j)-\overline{\Delta I} \right)^2,
\end{equation*}
where $\Omega(x, y)$ is a neighborhood around pixel $I(i, j)$, $\Delta I(i, j)$ is the Laplacian at pixel $I(i, j)$, and $\overline{\Delta I}$ is the average value of the image Laplacian within the pixel neighborhood $\Omega(x, y)$.
The Laplacian operator is effective in detecting blur due to its ability to measure regions with rapid intensity changes based on the second derivative of an image, similar to the Sobel and Scharr operators used for edge detection. 
This method assumes that an image with high variance contains both edge-like and non-edge-like features, indicating an in-focus image. 
Conversely, an image with low variance has a small spread of responses, indicating a lack of edges and bokeh style. 
Therefore, as blur increases, the number of distinct edges in an image decreases, leading to lower variance and a higher degree of blurriness.

\subsubsection{Brenner Score}
The Brenner score~\cite{Zahniser} is a measure of the focus quality of a digital image. 
The Brenner score computes image textures at two different scales to characterize the amount of high and low resolution data contained within the image. 
For example, texture measurements may be calculated from the average of adjacent pixel pairs from the high resolution measurements compared with the average of adjacent pixels triplets from the low resolution measurements. 
A score that indicated the quality of focus is then generated as a function of the low- and high-resolution measurements. 

\begin{figure}[h] 
\centering
\includegraphics[width=0.9\columnwidth]{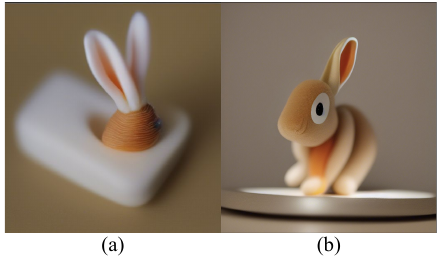}
\caption{
\textbf{Undesirable semantic mixing with a short global layout.}
We used a global prompt that incorporated both the terms \textit{``carrots"} and \textit{``rabbit"}. 
In the subsequent focus stage, a local prompt that solely included the keyword \textit{``carrots"} was added. 
When the global layout stage is too short (here, 20\%), the \textit{``carrots"} features merged with a \textit{``rabbit"}-like object, producing an object that combined the features of both for two distinct random seeds (a) and (b).
}
\label{Fig:warm-up}
\end{figure}

\begin{figure*}[h] 
\centering
\includegraphics[width=2.1\columnwidth]{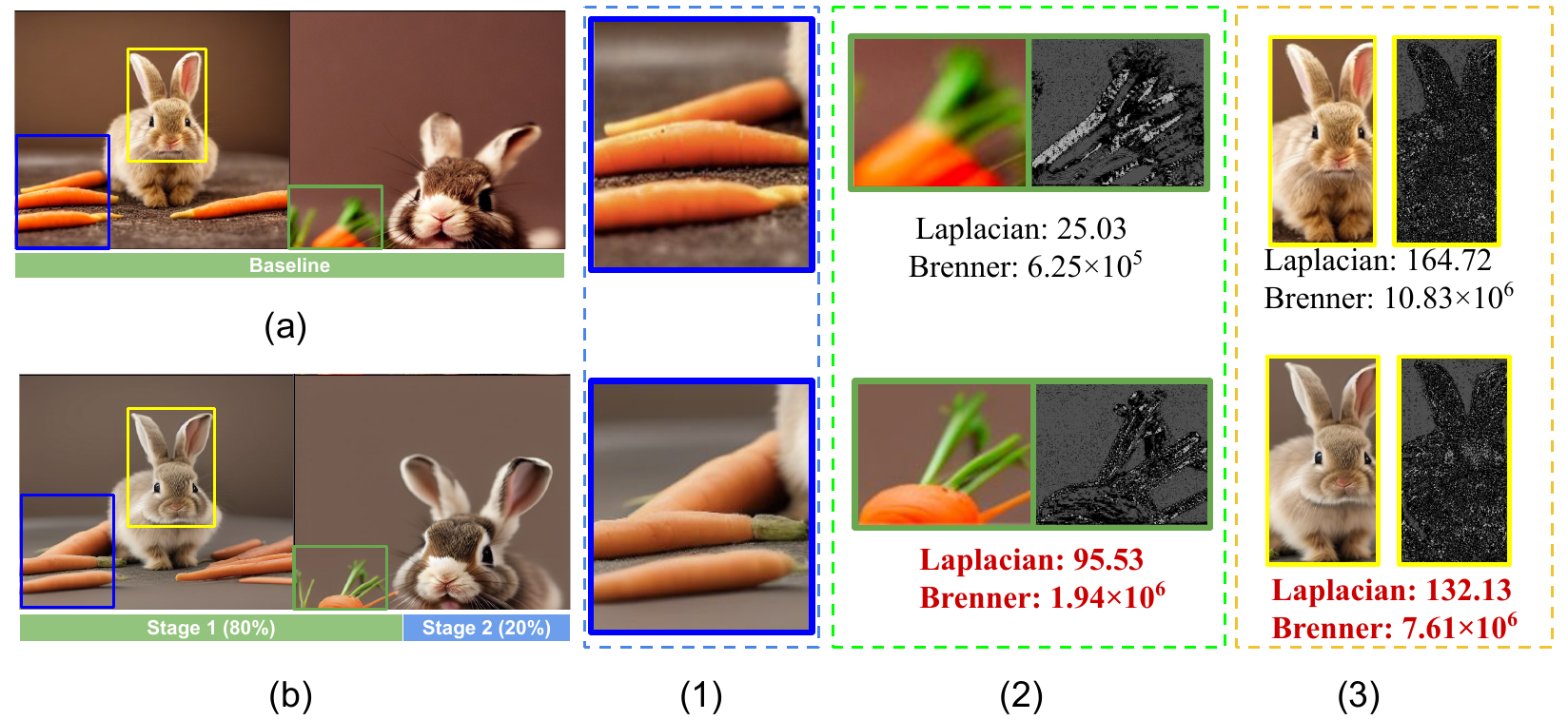}
\caption{\textbf{Comparison between the baseline LDM and \texttt{GBSD}.} 
We used the text conditioning \textit{``a cute baby bunny standing on top of a pile of baby carrots under a spot light"} both as the input prompt for LDM (a) and the global prompt for \texttt{GBSD} (b). 
In (b), a local prompt \textit{``a pile of baby carrots under a spot light"} was used in stage 2 with $80\%$ of denoising steps dedicated to stage 1 and $20\%$ to stage 2.
The highlighted image segments demonstrate that \texttt{GBSD} produces (1) stem details on the carrots, (2) a sharper image due to the focus on the carrots in stage 2, and (3) a bokeh effect on the bunny.
}
\label{Fig:change_stage}
\end{figure*}

\subsection{The Effect of Adjusting $\sigma$}
To investigate the impact of adjusting $\sigma$, which controls the proportion of the global layout (stage 1) and focus (stage 2) stages, we increased the length of the focus stage from 10\% to 50\% (Fig.~\ref{Fig:stage_comparison}).
We performed two experiments with varied $\sigma$ both containing the same global prompt: \textit{``a cute baby bunny standing on top of a pile of baby carrots under a spot light"}. 
The first experiment (Fig.~\ref{Fig:stage_comparison}, top row) used a local prompt for the focus stage of \textit{``a pile of carrots"}, whereas the second experiment (Fig.~\ref{Fig:stage_comparison}, bottom row) used the local prompt \textit{``a cute rabbit"}.

Firstly, both \textit{``rabbit"} and \textit{``carrots"} objects are easily identifiable when the global layout stage encompasses at least 70\% of the total denoising process (Fig.~\ref{Fig:stage_comparison} (a-c)), indicating that a stage 1 length of 70\%-90\% of the denoising steps is sufficient to stabilize objects.
When the global layout stage was allocated 60\% or less of the total denoising process, there existed insufficient evidence to confirm the presence of the feature that was the target of the bokeh effect, i.e., the \textit{``rabbit''} in Figure~\ref{Fig:stage_comparison}, top row (d-e) and the \textit{``carrots''} in Figure~\ref{Fig:stage_comparison}, bottom row (d-e).
Furthermore, by comparing the different $\sigma$ within each experiment, it can be observed that an increase in stage 2 leads to a degradation in the information that is present in the global prompt but not in the local prompt.
When comparing the two experiments for each $\sigma$, we observed that the object position within each image are essentially identical since we use the same configuration to generate the images. 
However, the preservation of the object features are reversed with the first experiment retaining the features of \textit{``carrots''} (Fig.~\ref{Fig:stage_comparison}, top row) and second experiment retaining the features of \textit{``rabbit"} (Fig.~\ref{Fig:stage_comparison}, bottom row).

We also investigated the behavior of stage diffusion when the global layout stage was shortened to 20\% of the denoising process (Fig.~\ref{Fig:warm-up}). 
Regardless of the local prompt, a short global layout stage produced objects that semantically intermixed the features of \textit{``carrots"} and \textit{``rabbit"} as a result of not being able to appropriately establish image layout before refining details.

\begin{figure*}[htbp] 
\centering
\includegraphics[width=2.1\columnwidth]{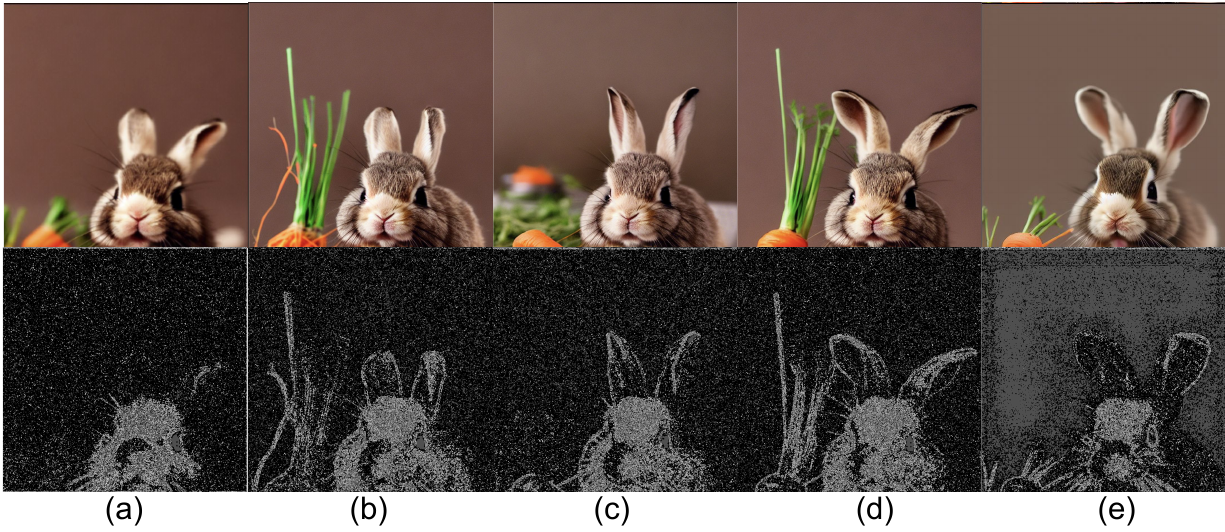} 
\caption{\textbf{Comparisons with prompt containing keyword \textit{``focus"}.}
We implemented variable focus using different prompts for LDM (a-d) and the same prompt as in Figure~\ref{Fig:change_stage} for \text{GBSD} (e):  (a) \textit{``a cute baby bunny standing on top of a pile of baby carrots under a spot light"}, (b) \textit{``a cute baby bunny standing on top of a pile of baby carrots under a spot light, \textbf{rabbit is in focus}, \textbf{carrots out of focus}"}, (c)\textit{``a cute baby bunny standing on top of a pile of baby carrots under a spot light, \textbf{carrots is in focus}, \textbf{rabbit out of focus}"}, (d)\textit{``a cute baby bunny standing on top of a pile of baby carrots under a spot light, \textbf{rabbit and carrot in focus}"} and (e) our output from Figure~\ref{Fig:change_stage}.}
\label{Fig:focus}
\end{figure*}

\subsection{Baseline Comparisons}
We quantitatively compared the generated images between the baseline LDM~\cite{Rombach_2022_CVPR_ldm} and \texttt{GBSD} (Fig.~\ref{Fig:change_stage}) using the VoL and Brenner measure.
When a bokeh effect is desired, the VoL and Brenner measure should be lower; conversely, when sharp details are desires, the VoL and Brenner measure should be larger.
Using LDM and \texttt{GBSD}, we generated two images using two distinct random seeds (Fig.~\ref{Fig:change_stage}, (a) and (b), left and right).
We used the prompt \textit{``a cute baby bunny standing on top of a pile of baby carrots under a spot light"} as an input prompt for the baseline and global prompt for \texttt{GBSD}. 
We used a partial segment of the global prompt as the local prompt, \textit{``a pile of baby carrots under a spot light"}, which encompassed $20\%$ of denoising steps for \texttt{GBSD} in the focus stage.

Even when the random input seed is fixed across models, some segments of the image were too dissimilar to produce meaningful quantitative comparisons (Fig.~\ref{Fig:change_stage} (1)).
However, even these segments of carrots show enhanced detail in the \texttt{GBSD} synthesized image compared with LDM (e.g., the green stem).
Next, we highlighted the highly similar carrot objects in the second experiment (Fig.~\ref{Fig:change_stage}, green boxes, right image) and computed their corresponding blur maps (Laplacian of the image normalized to grayscale).
While the normalization of the blur map makes it difficult to compare the LDM and \texttt{GBSD}, the VoL, Brenner score, and qualitative comparisons demonstrate that the carrots are more in-focus for \texttt{GBSD} (Fig.~\ref{Fig:change_stage} (2)) where our approach achieves 95.53 in Laplacian score and 1.94$\times$10$^{6}$ in Brenner score which is 3.81$\times$ and 3.10$\times$ to baseline.
When a bokeh effect is desired (here, in the bunny), the \texttt{GBSD} produces smaller (better) VoL and Brenner score values (Fig.~\ref{Fig:change_stage} (3)).



\subsection{Comparison with \textit{``focus"} Prompt}
As discussed in the previous section, our proposed method demonstrated a significant score improvement in terms of Laplacian measures compared to the baseline approach when evaluated with the same prompt, that is, the LDM prompt and \texttt{GBSD} global prompts were identical. 
Furthermore, the focus prompt did not contain any additional information.
We investigated whether adding text to the conditioning prompt could reproduce sharpening or blurring effects, we added \textit{``in focus"} and \textit{``out of focus"} phrases to the prompts (Fig.~\ref{Fig:focus}).


Using the same seed, we synthesized $4$ images using LDM (Fig.~\ref{Fig:focus} (a-d)).
Image (a) was generated using the prompt \textit{``a cute baby bunny standing on top of a pile of baby carrots under a spot light"}. 
To generate images (b-d), we added suffixes to change the image focus: (b) ``\textbf{rabbit is in focus}, \textbf{carrots out of focus}''; (c) ``\textbf{carrots is in focus}, \textbf{rabbit out of focus}''; (d) ``\textbf{rabbit and carrot in focus}''. 
Finally, image (e) was generated using stage diffusion as described above (Fig.~\ref{Fig:change_stage}). 
Visual inspection of the images demonstrate undesirable results for images (a-c): (1) a bokeh effect in the carrots for image (a) when it was not desired; (2) an unnatural carrot structure being created in image (b) in response to the \textit{``out of focus''} phrase; and (3) the opposite effect of what was desired in image (c), that is, the carrots are blurry then they should be in focus.
Conversely, image (d) shows both rabbit and carrots in focus which, while desired based on the text prompt, does not achieve a bokeh effect.
However, compared with the LDM baseline, the desired sharpening of the carrots and bokeh effect on the bunny is achieved in image (e) with \texttt{GBSD} (see also Fig.~\ref{Fig:change_stage}).

\begin{figure}[!htbp]
\centering
\includegraphics[width=1.05\columnwidth]{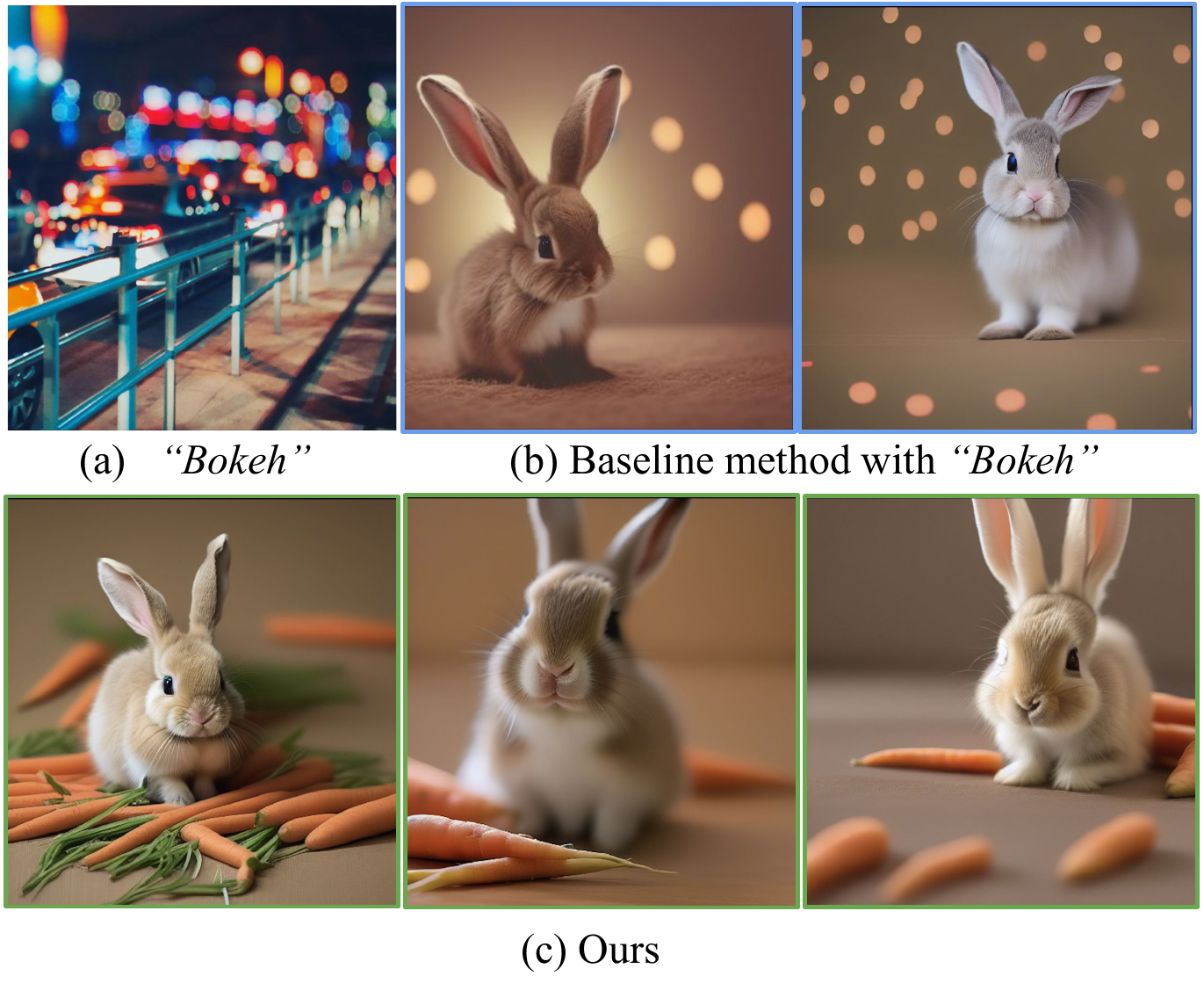}
\caption{\textbf{Comparing the \textit{``bokeh effect"} between LDM with \textit{``bokeh"} in the prompt and \texttt{GBSD}.} 
(a) is an example of \textit{``bokeh"} from a photography resource~\cite{phlearn_2020}. (b) is the output of LDM where the \textit{``bokeh"} keyword is added to the text prompt. (c) is the output of \texttt{GBSD} with a focus stage prompt highlighting the bunny (left and right images) and carrots (middle image).}



\label{Fig:Bokeh}
\end{figure}

\begin{figure*}[h!] 
\centering
\includegraphics[width=2\columnwidth]{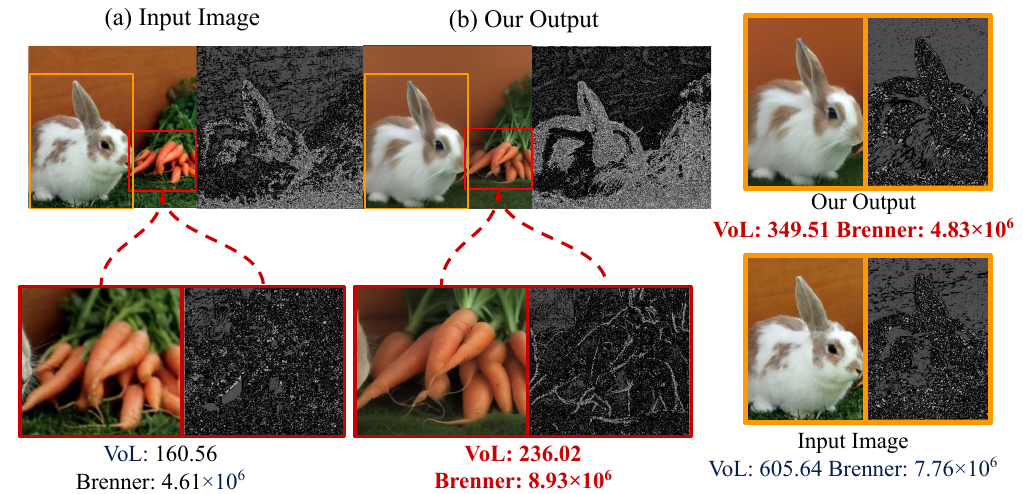}
\caption{\textbf{Evaluating image-to-image generation.}
The global prompt was ``\textit{A cute rabbit stands with carrots with green leaf}" and the local prompt is ``\textit{carrots with green leaf}" for (b). 
Our outputs achieve up to 73$\%$ improvement in VoL and 93$\%$ improvement in Brenner compared to the input image.}
\label{Fig:shift focus}
\end{figure*}

\subsection{Comparisons with \textit{``bokeh"} Prompt}


Next, we investigated whether adding a \textit{``bokeh effect"} prompt to the baseline would produce the desired blurring of semantically distinct objects (Fig.~\ref{Fig:Bokeh}).
Typical examples of the bokeh effect includes urban settings with a blurring ball out-of-focus effect on lighting in the background  (Fig.~\ref{Fig:Bokeh} (a)).
Including the keywords \textit{``bokeh effect"} in the prompt of LDM interestingly removed carrot objects but generated a similar blurring effect with carrot-colored lighting (Fig.~\ref{Fig:Bokeh} (b)). 
The bokeh effect produces by LDM is also not realistic, as there are no physical light sources located in the background.
Since the presence of points of light is a common feature in publicly available bokeh images, the bokeh artifact produced by LDM is likely due to its presence in the training data. 
In contrast, \texttt{GBSD} offers the advantage of producing a more realistic bokeh effect on either carrot or bunny objects instead of simply replicating a typical bokeh effect from the training data (Fig.~\ref{Fig:Bokeh} (c)). 

\subsection{Focus Shift in Image-to-Image Generation}
Text-to-image generation involves generating an image from a text prompt, whereas image-to-image generation requires two types of conditioning: a text prompt and an input image. 
One challenge of image-to-image generation is identifying an appropriate level of noise to incorporate into the input image. 
The noise must be capable of modifying image features without causing significant deviations from the input image.
An image was selected from an online source to use as the input (Fig.~\ref{Fig:shift focus} (a)).
To add focus to the carrots, we set a global prompt ``\textit{A cute rabbit stands with carrots with green leaf}" in the global layout stage and local prompt of ``\textit{carrots with green leaf}" in focus stage. 
Overall, \texttt{GBSD} sharpens the detail of the carrots achieving a VoL of 236.02 and Brenner score of 8.93$\times$ 10$^{6}$, which are 1.73$\times$ and 1.93$\times$ larger than the input image, respectively (Fig.~\ref{Fig:shift focus} (b), left). 
Further, the stage diffusion algorithm also achieves a bokeh effect on the rabbit, with a smaller VoL and Brenner score when compared with the input image (Fig.~\ref{Fig:shift focus} (b), right).



\section{Conclusions}
In this paper, we presented \texttt{GBSD}, the first generative text-to-image model that synthesizes photorealistic images with a bokeh style. 
The approach combines latent diffusion models with a 2-stage conditioning algorithm to render blurring effects. 
Unlike prior bokeh methods, \texttt{GBSD} is able to produce a semantic bokeh effect, where semantically distinct objects are blurred based on the 2-stage text conditioning procedure.
We evaluated \texttt{GBSD} both quantitatively and qualitatively and demonstrated its ability to be applied in both text-to-image and image-to-image settings. 
In sum, we believe that \texttt{GBSD} and other generative models of photorealistic images with artistic stylings can provide a valuable content generation resource to AI-assisted industries reliant on image synthesis. 

\clearpage

{\small
\bibliographystyle{ieee_fullname}
\bibliography{ref}
}

\end{document}